\documentclass[letterpaper]{article}

\usepackage{helvet}
\usepackage{courier}
\usepackage{multirow}
\usepackage{amsmath}
\usepackage{graphicx}

\makeatletter

\pdfpageheight\paperheight
\pdfpagewidth\paperwidth

\providecommand{\tabularnewline}{\\}

\def\year{2016}
\usepackage{aaai16}
\usepackage{times}

\usepackage{subcaption}

\usepackage{algorithm}
\usepackage{algorithmic}
\renewcommand{\algorithmicrequire}{ \textbf{Input:}}
\renewcommand{\algorithmicensure}{ \textbf{Output:}}
\makeatother
\newcommand\Mark[1]{\textsuperscript#1}

\begin{document}

\title{Learning to Generate Posters of Scientific Papers\thanks{This work is supported by NSFC (61333014, 61373059, and 61321491) and JiangsuSF (BK20150016).}}

\author{Yuting Qiang\Mark{1}, Yanwei Fu\Mark{2}, Yanwen Guo\Mark{1}\thanks{Corresponding author}, Zhi-Hua Zhou\Mark{1} \and Leonid Sigal\Mark{2}\\
\Mark{1} National Key Laboratory for Novel Software Technology, Nanjing University, Nanjing 210023, China \\
\Mark{2} Disney Research Pittsburgh, 4720 Frobes Avenue, Lower Level, 15213, USA \\
\{qiangyuting.new,ywguo.nju\}@gmail.com, zhouzh@nju.edu.cn, \{yanwei.fu,lsigal\}@disneyresearch.com
}

\maketitle
\begin{abstract}
\begin{quote}
Researchers often summarize their work in the form of posters. Posters provide a coherent and efficient way to convey core ideas from scientific papers.
Generating a good scientific poster, however, is a complex and time consuming cognitive task, since such posters need to be {\emph{readable}}, {\emph{informative}}, and visually {\emph{aesthetic}}.
In this paper, for the first time, we study the challenging problem of learning to generate posters from scientific papers.
To this end, a data-driven framework, that utilizes graphical models, is proposed.
Specifically,  given content to display, the key elements of a good poster, including panel layout and attributes of each panel, are learned and inferred from data. Then, given inferred layout and attributes, composition of graphical elements within each panel is synthesized.
To learn and validate our model, we collect and make public a Poster-Paper dataset, which consists of scientific papers and corresponding posters with exhaustively labelled panels and attributes.
Qualitative and quantitative results indicate the effectiveness of our approach.

\end{quote}
\end{abstract}

\section{Introduction}

The emergence of large number of scientific papers in various academic fields and venues (conferences and journals) is noteworthy.
For example, IEEE Conference on Computer Vision and Pattern Recognition (CVPR) accepted over 600 papers in 2016 alone.
It is time-consuming to read all of these papers for the researchers, particularly those interested to holistically assess state-of-the-art or emerge with understanding of core scientific ideas explored in the last year.
Converting a conference paper into a poster provides important means to efficiently and coherently convey core ideas and findings of the original paper.
To achieve this goal, it is therefore essential to keep the posters readable, informative and visually aesthetic.
It is challenging, however, to design a high-quality scientific poster which meets all of the above design constraints,
particularly for those researchers who may not be proficient at design tasks or familiar with design packages (e.g., Adobe Illustrator).

In general, poster design is a complicated and time-consuming task; both understanding of the paper content and experience in design are required.

Automatic tools for scientific poster generation would help researchers by providing them with an easier way to effectively share their research. Further, given avid amount of scientific papers on ArXiv and other on-line repositories, such tools may also provide a way for other researchers to consume the content more easily. Rather than browsing raw papers, they may be able to browse automatically generated poster previews (potentially constructed with their specific preferences in mind).

However, in order to generate a scientific poster in accordance with, and representative of, the original paper, many problems need to be solved:
1) {\em Content extraction.} Both important textual and graphical content needs to be extracted from the original paper;
2) {\em Panel layout.} Content should fit each panel, and the shape and position of panels should be optimized for readability and design appeal;
3) {\em Graphical element (figures and tables) arrangement.} Within each panel, textual content can typically be sequentially itemized, but for graphical elements, their size and placement should be carefully considered.
Due to these challenges, there are few automatic tools for scientific poster generation.

In this paper, we propose a data-driven method for automatic scientific poster generation (given a corresponding paper). Contents extraction and layout generation are two key components in this process. For content extraction, we use TextRank \cite{MihalceaT04emnlp} to extract textual content, and provide an interface for extraction of graphical content (e.g., figures, tables, etc.). Our approach focuses primarily on poster layout generation.
We address the layout in three steps. First, we propose a simple probabilistic graphical model to infer
panel attributes. Second, we introduce a tree structure to represent panel layout, based on which we further design a recursive algorithm to generate new layouts. Third, in order to synthesize layout within each panel, we train another probabilistic graphical model to infer the attributes of the graphical elements.

Compared with posters designed by the authors, our approach can generate different results to adapt to different paper sizes/aspect ratios or styles, by training our model with different dataset, and thus provides more expressiveness in poster layout.
To the best of our knowledge, this paper presents the first framework for poster generation
from the original scientific paper. \\
\indent Our paper makes the following contributions:
\begin{itemize}
\item Probabilistic graphical models are proposed to learn scientific poster design patterns, including panel attributes and graphical element attributes, from existing posters.
\item A new algorithm, that considers both information conveyed and aesthetics, is developed to generate the poster layout.
\item We also collected and make available a Poster-Paper dataset with labelled poster panels and attributes.
\end{itemize}

\section{Related Work}
\noindent \textbf{General Graphical Design.} Graphical design has been studied extensively in computer graphics community. This involves several related, yet different topics, including \emph{text-based layout generation} \cite{Jacobs2003ACM,Damera2011,Hurst2009}, \emph{single-page graphical design} \cite{odonovan2014,harrington2004aesthetic}, \emph{photo albums layout} \cite{Geigel03Mul}, \emph{furniture layout} \cite{Merrell11ACM,craigyu2011furniture}, and even \emph{interface design} \cite{Gajos05UIST}.
Among them, text-based layout pays more attention on informativeness, while attractiveness also needs to be considered in poster generation. Other topics would take aesthetics as the highest priority. However, some principles (such as alignment or read-order) need to be followed in poster design. In summary, poster generation needs to consider readability, informativeness and aesthetics of the generated posters simultaneously.

\noindent \textbf{Manga Layout Generation.} Several techniques have been studied to facilitate layout generation for western comics or manga. For example,, for example, \emph{scene frame extraction} \cite{arai2010method,Pang2014MM}, \emph{automatic stylistic manga layout generation} \cite{Cao2012TOG,Jing15TMM}, and \emph{graphical elements composition} \cite{Cao2014Siggraph}. For preview generation of comic episodes \cite{Hoashi2011MM}, both frame extraction and layout generation are considered. Other research areas, such as \emph{manga retargeting} \cite{Matsui2011SIG} and \emph{manga-like rendering} \cite{Qu2008ACM} also draw considerable attention. However, none of these methods can be directly used to generate scientific posters, which is the focus of this paper.\par
Our panel layout generation is inspired by the recent work on Manga layout \cite{Cao2012TOG}. We use a binary tree to represent the panel layout. By contrast, the manga Layout trains a Dirichlet distribution to sample a splitting configuration, and different Dirichlet distribution for each kind of instance need to be trained. Instead, we propose a recursive algorithm to search for the best splitting configuration along a tree.

\section{Overview}
\textbf{Problem Formulation.} Assume that we have a set of posters \textbf{M} and their corresponding scientific papers. Each poster $m\in\mathbf{M}$ includes a set of panels $\mathbf{P}_{m}$, and each panel $p\in\mathbf{P}_{m}$ has a set of graphical elements (figures and tables) $\mathbf{G}_{p}$. Each panel $p$ is characterized by five attributes:
\begin{description}
\item [{text~length~($l_{p}$)}] text length within a panel;
\item [{text~ratio~($t_{p}$)}] text length within a panel relative to text length of the whole poster, $t_{p}=l_{p}/\sum_{q\in\mathbf{P}_{m}}l_{q}$;
\item [{graphical~elements~ratio~($g_{p}$)}]\footnote{Note that there is a little difference between this variable and text ratio $t_{p}$. We do not use the figure size in poster. Instead, we use the corresponding figure from the original paper.} the size of graphical elements within a panel relative to the total size of graphical elements in the poster.
\item [{panel~size~($s_{p}$)~and~aspect~ratio~($r_{p}$),}] $s_{p}=w_{p}\times h_{p}$ and $r_{p}=w_{p}/h_{p}$, where ${w_{p}}$ and ${h_{p}}$ denote the width and height of a panel with respect to the poster, separately.
\end{description}
Each graphical element $g\in\mathbf{G}_{p}$ has four attributes:
\begin{description}
\item [{graphical~element~size~($s_{g}$)} and aspect ratio ($r_{g}$),] $s_{g}=w_{g}\times h_{g}$ and $r_{g}=w_{g}/h_{g}$, where ${w_{g}}$ and ${h_{g}}$ denote the width and height of a graphical element relative to the whole paper respectively;
\item [{horizontal~position~($h_{g}$)}] we assume that panel content is arranged sequentially from top to bottom\footnote{This holds true when using latex beamer to make posters.}; hence only relative horizontal position needs to be  considered, which is defined by a discrete variable $h_{g} \in \{left,center,right\}$;
\item [{graphical~element~size~in~poster~($u_{g}$)}] is the ratio of the width of the graphical element with width of the panel.
\end{description}
To learn how to generate the poster, our goal is to \emph{determine the above attributes of each panel $p$ and each graphical element $g\in\mathbf{G}_{p}$, as well as to infer the arrangement of all panels.}

Intuitively, a trivial solution is to use a learning model (e.g., SVR) to learn how to regress these attributes, including $s_{p}$, $r_{p}$,
$u_{g}$, and $h_{g}$, while regarding $t_{p}$, $g_{p}$ , $l_{p}$, $r_{g}$, and $s_{g}$ as features. However, such a solution lacks an insight mechanism for exploring the relationships between the panel attributes (e.g., $s_{p}$) and graphical elements attributes (e.g., $u_{g}$). And it may fail to meet the requirements of readability, informativeness, and aesthetics. We thus propose a novel framework to solve our problem.

\begin{figure*}
\centering{} \includegraphics[width=0.8\textwidth]{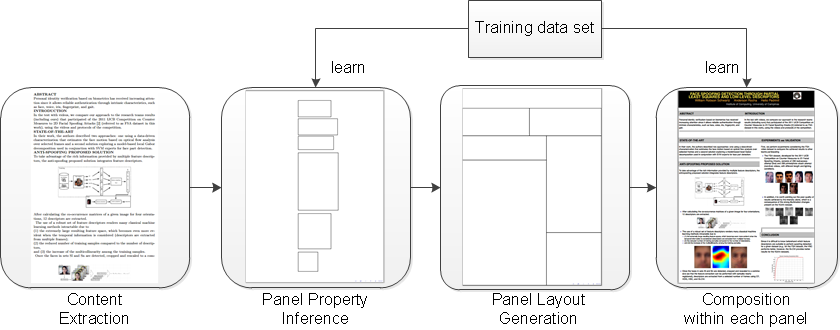}
\protect\protect\protect\caption{\label{fig:Overview-of-our} Overview of the proposed approach.}
\end{figure*}

\noindent \textbf{Overview.} To generate a \emph{readable}, \emph{informative} and \emph{aesthetic} poster, we simulate the rule-of-thumb on how people design the posters in practice. We generate the panel layout, then arrange the textual and graphical elements within each panel.\par

Our framework overall has four steps (as shown in Figure~\ref{fig:Overview-of-our}). However, the core of our framework focuses on three specific algorithms designed to facilitate poster generation.
We first extract textual content from the paper using TextRank~\cite{MihalceaT04emnlp}%
\footnote{We use TextRank for text content extraction, however, TextRank can be replaced with other state-of-the-art textual summary algorithms. %
}, this will be detailed in the Experimental Result section. Non-textual content (figures and tables) are extracted by user interaction. All these extracted contents are sequentially arranged and represented by the first blob in Figure~\ref{fig:Overview-of-our}. Inference of the initial panel key attributes (such as panel size $s_{p}$ and aspect ratio $r_{p}$) is then conducted by learning a probabilistic graphical model from the training data. Furthermore, \emph{panel layout} is synthesized by developing a recursive algorithm to further update these key attributes (i.e., $s_{p}$ and $r_{p}$) and generate an \emph{informative} and \emph{aesthetic} panel layout. Finally, we compose panels by utilizing the graphical model to further synthesize the visual properties of each panel (such as the size and position of its graphical elements).

\section{Methodology}

\noindent \textbf{Panel Attribute Inference.} Our approach tries to divide a scientific poster into several rectangular panel blocks. Each panel should not only be of an appropriate size, to contain corresponding textual and graphical content, but also be in a suitable shape (aspect ratio) to maximize aesthetic appeal. Our approach learns a probabilistic graphical model to infer the initial values for the size and aspect ratio of each panel.

As each panel is composed of both textual description and graphical elements, we assume that panel size ($s_{p}$) and aspect ratio ($r_{p}$) are conditionally dependent on text ratio $t_{p}$ and graphical element ratio $g_{p}$. Therefore, the likelihood of a set of panels $p$ can be defined as:
\begin{equation}
Pr(s_{p},r_{p}|t_{p},g_{p})=\prod_{p\in P}Pr(s_{p}|t_{p},g_{p})Pr(r_{p}|t_{p},g_{p})\label{eq:panel_infer}
\end{equation}
where $Pr(s_{p}|t_{p},g_{p})$ and $Pr(r_{p}|t_{p},g_{p})$ are conditional probability distributions (CPDs) of $s_{p}$ and $r_{p}$ given $t_{p}$
and $g_{p}$. We define them as two conditional linear Gaussian distributions:
\begin{equation}
Pr(s_{p}|t_{p},g_{p})=N({s_{p}};\mathbf{w_{s}}\cdot[t_{p},g_{p}, 1]^\mathsf{T},\mathbf{\sigma_{s}})
\end{equation}
\begin{equation}
Pr(r_{p}|t_{p},g_{p})=N({r_{p}};\mathbf{w_{r}}\cdot[t_{p},g_{p}, 1]^\mathsf{T},\mathbf{\sigma_{r}})
\end{equation}
where $t_{p}$ and $g_{p}$ are defined by the \emph{content extraction} step demonstrated in Figure \ref{fig:Overview-of-our}; $\mathbf{w_{s}}$ and $\mathbf{w_{r}}$ are the parameters that leverage the influence of various factors; $\mathbf{\sigma_{s}}$ and $\mathbf{\sigma_{r}}$ are the variances. The parameters ($\mathbf{w_{s}}$, $\mathbf{w_{r}}$, $\mathbf{\sigma_{s}}$ and $\mathbf{\sigma_{r}}$) are estimated using maximum likelihood from training data.
Using the learned parameters, initial attributes of each panel can be inferred.

Note that in order to learn from limited data, this step actually employs two assumptions:
(1) $s_{p}$ and $r_{p}$ are conditionally independent;
(2) The attribute sets of panels are independent.
We need the panels to be neither too small in size ($s_p$), nor too distorted in aspect ratio ($r_p$), to ensure readable, informative and aesthetic poster. The two assumptions introduced here are sufficient for this task.
Furthermore, the attribute values estimated from this step are just good initial values for the property of each panel. We use the next two steps to further relax these assumptions and discuss the relationship between $s_{p}$ and $r_{p}$, as well as the relationship among different panels (Algorithm \ref{alg:panel_gen}).

To ease exposition, we denote the set of panels as $L=\{(s_{p_{1}},r_{p_{2}}), (s_{p_{2}},r_{p_{2}}), \cdots, (s_{p_{k}},r_{p_{k}})\}$, where $s_{p_{i}}$ and $r_{p_{i}}$ are the size and aspect ratio of $i$th panel $p_{i}$, separately; with $\left|L\right|=k$.\par

\begin{figure} \centering
\begin{subfigure}[t]{0.2\textwidth}
\includegraphics[width=\textwidth]{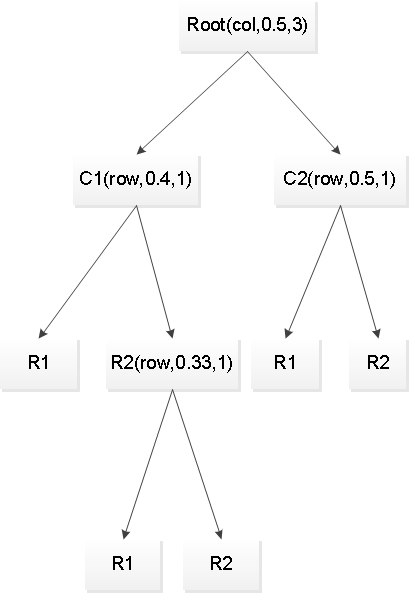}
\end{subfigure}
\begin{subfigure}[t]{0.2\textwidth}
\includegraphics[width=\textwidth]{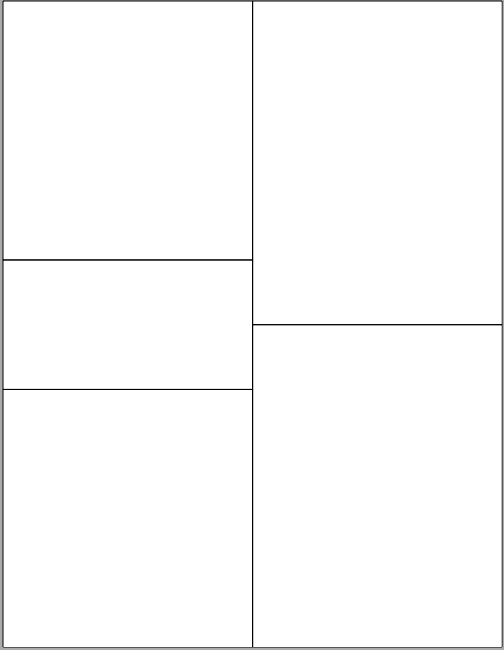}
\end{subfigure}
\caption{\label{fig:tree_structure}Panel layout and the corresponding tree structure.  The tree structure of a poster layout contains five panels. The first splitting is vertical with the splitting ratio  (0.5, 0.5). The poster is further divided into three panels in the left, and two panels in the right. This makes the whole page as two equal columns. For the left column,  we resort to a horizontal splitting with the splitting ratio (0.4, 0.6). The larger one is further horizontally divided into two panels with the splitting ratio (0.33, 0.67). We only split the right column once, with the splitting ratio (0.5, 0.5).}
\end{figure}

\noindent\textbf{Panel Layout Generation}. One conventional way to design posters is to simply arrange them in two or three columns style. This scheme, although simple, however, makes all posters look similar and unattractive. Inspired by manga layout generation~\cite{Cao2012TOG}, we propose a more vivid panel layout generation method. Specifically, we arrange the panels with a binary tree structure to help represent the panel layout. The panel layout generation is then formulated as a process of recursively splitting of a page, as is illustrated and explained in Figure \ref{fig:tree_structure}.\par
Conveying information is the most important goal for a scientific poster, thus we attempt to maintain relative size for each panel during panel layout generation. This motivates the following loss function for the panel shape variation,

\begin{equation}
l(p_{i})=|r_{p_{i}}-r_{p_{i}}^{'}|
\end{equation}
where $r_{p_{i}}^{'}$ is the aspect ratio of a panel after optimization. This will lead to a combined aesthetic loss for the poster,
\begin{equation}
Loss(L,L^{'})=\sum_{i=1}^{k}{l(p_{i})}\label{eq:panel_variation}
\end{equation}
where $L^{'}$ is the poster panel set after optimization. In each splitting step, the combinatorial choices for splitting positions can be recursively computed and compared with respect to the loss function above. We choose the panel attributes with the lowest loss (Eq. \ref{eq:panel_variation}). The whole algorithm is summarized in Algorithm \ref{alg:panel_gen}.

\begin{algorithm}[htb]
\caption{Panel layout generation} \label{alg:panel_gen}
\begin{algorithmic}[1] \REQUIRE ~~\\
Panels which we learned from graphical model\\
$L=\{(s_{p_1},r_{p_1}),(s_{p_2},r_{p_2}), \cdots,(s_{p_k},r_{p_k})\}$;\\
rectangular page area $x$, $y$, $w$, $h$. \\ \ENSURE ~~\\
\IF {$k==1$}
\STATE adjust panels$[0]$ to adapt to the whole rectangular page area, return the aesthetic loss: $|r_{p_0}-w/h|$;
\ELSE
\FOR {each $i\in[1,k-1]$}
\STATE $t=\sum_{j=1}^{i} s_{p_j} / \sum_{j=1}^{n} s_{p_j}$;
\STATE $Loss_1$ = Panel Arrangement($(s_{p_1},r_{p_1}),(s_{p_2},r_{p_2}),$ \\
$\cdots,(s_{p_i},r_{p_i})$, $x$, $y$, $w$, $h\times t$); \STATE $Loss_2$ = Panel Arrangement($(s_{p_{i+1}},r_{p_{i+1}}),(s_{p_{i+2}},$\\ $r_{p_{i+2}}), \cdots,(s_{p_k},r_{p_k})$, $x$, $y+h\times t$, $w$, $h\times(1-t)$);
\IF {$Loss > Loss_1+Loss_2$} \STATE $Loss = Loss_1+Loss_2$; \STATE record this arrangement;
\ENDIF
\STATE $Loss_1$ = Panel Arrangement($(s_{p_1},r_{p_1}),(s_{p_2},r_{p_2}),$\\
$\cdots,(s_{p_i},r_{p_i})$, $x$, $y$, $w\times t$, $h$);
\STATE $Loss_2$ = Panel Arrangement($(s_{p_{i+1}},r_{p_{i+1}}),(s_{p_{i+2}},$\\
$r_{p_{i+2}}),\cdots,(s_{p_k},r_{p_k})$, $x+w*t$, $y$, $w\times(1-t)$, $h$);
\IF {$Loss > Loss_1+Loss_2$} \STATE $Loss = Loss_1+Loss_2$; \STATE record this arrangement;
\ENDIF
\ENDFOR
\ENDIF
\RETURN Loss and arrangement.

\end{algorithmic}
\end{algorithm}

\noindent \textbf{Composition within a Panel.}
Having inferred layout of the panels, we turn our attention to composition of graphical elements within the panels.
We model and infer attributes of graphical elements using another probabilistic graphical model. Particularly, the key attributes we need to estimate are the horizontal position $h_{g}$ and graphical element size $u_{g}$. In our model, horizontal position $h_{g}$ relies on $s_{p}$, $l_{p}$ and $s_{g}$, while $u_{g}$ relies on $r_{p}$, $s_{g}$ and $r_{g}$, so the likelihood is
\begin{equation}
\begin{split}Pr(h_{g},u_{g}| & s_{p},r_{p},l_{p},s_{g},r_{g})=\\
 & \prod_{p\in P}\prod_{g\in p}Pr(h_{g}|s_{p},l_{p},s_{g})Pr(u_{g}|r_{p},s_{g},r_{g})
\end{split}
\label{equ:likelihood_function}
\end{equation}
$Pr(u_{g}|s_{p},l_{p},s_{g})$ and $Pr(h_{g}|r_{p},s_{g},r_{g})$ are the conditional probability distributions (CPDs) of $u_{g}$ and
$h_{g}$ given $s_{p},l_{p},r_{p},s_{g}$ and $r_{g}$ respectively. The conditional linear Gaussian distribution is also used here,

\begin{equation}
Pr(u_{g}|s_{p},l_{p},s_{g})=N({u_{g}}|\mathbf{w_{u}}\cdot[s_{p},l_{p},s_{g},1]^\mathsf{T},\mathbf{\sigma_{u}})
\end{equation}
where $\mathbf{w_{u}}$ is the parameter to balance the influence of different factors. Since we take horizontal position $h_{g}$ as an enumerated variable, a natural way to estimate it is to make it a classification problem by using the softmax function,
\begin{equation}
Pr(h_{g}=i|r_{p},s_{g},r_{g})=\frac{e^{\mathbf{w_h}_{i}\cdot[r_{p},s_{g},r_{g},1]^{\mathsf{T}}}}{\sum_{j=1}^{H}e^{\mathbf{w_h}_{j}\cdot[r_{p},s_{g},r_{g},1]^{\mathsf{T}}}}
\end{equation}
where $H$ is the cardinality of the value set of $h_{g}$, i.e. $H=3$, $\mathbf{w_h}_{i}$ is the $i$th row of $\mathbf{w_h}$. The maximum likelihood method is used to estimate parameters, including $\mathbf{w}_{\mathbf{u}}$,$\mathbf{w_{h}}$ and $\sigma_{\mathbf{u}}$.

Different from Eq. \ref{eq:panel_infer}, directly inferring $h_{g}$ and $u_{g}$ is not advisable, since the panel content may exceed the panel bounding box and affect the aesthetic measure of a poster. To avoid this problem, we employ the likelihood-weighted sampling method \cite{weighted_sampling} to generate samples from the model,
by maximizing the likelihood function (Eq. \ref{equ:likelihood_function}) with this strict constraint,
\begin{equation}
{\sum_{g\in p}h_{p} \times u_{g}}+{\alpha\times\beta\times l_{p}/w_{p}}<h_{p}\label{eq:constraint}
\end{equation}
where $\alpha$ and $\beta$ denote the width and height of a single character respectively. The first term of the above constraint indicates the height of graphical elements while the second term represents the height of textual contents.

\section{Experimental Results}

\noindent \textbf{Experimental Setup.}
We collect and make available to the community the
first Poster-Paper dataset.
Specifically, we selected $25$ well-designed pairs of scientific papers and their corresponding posters from $600$ publicly available pairs we collected. These papers are all about scientific topics, and their posters have relatively similar design styles. We  further annotate panel attributes, such as panel width, panel height and so on. We make a training and testing split: $20$ pairs for training and five for testing. There is total of $173$ panels in our dataset. $143$ for training and $30$ for testing.

We use TextRank to extract textual content from the original paper. In order to give different importance of different sections, we can set different extraction ratio for each section. This will result in important sections generating more content and hence occupying bigger panels. For simplicity, this paper uses equal important weights for all sections. User-interaction is also required to highlight and select important figures and tables from original paper. We use the Bayesian Network Toolbox (BNT)~\cite{Murphy02} to estimate key parameters. For graphical element attributes inference, we generate $1000$ samples by the likelihood-weighted sampling method \cite{weighted_sampling} for Eq. \ref{equ:likelihood_function} while the constraint Eq.\ref{eq:constraint} is used. With the inferred metadata, the final poster is generated in latex Beamerposter format with Lankton theme.

For baseline comparison, we invite three second-year Phd students, who are not familiar with our project, to hand design posters for the test set. These three students work in computer vision and machine learning and have not yet published any papers on these topics; hence they are novices to research. Given the test set papers, we ask the students to work together and design a poster for each paper.

\noindent\textbf{Running time.} Our framework is very efficient. Our experiments were done on a PC with an Intel Xeon 2.0 GHz CPU and 144GB RAM. Tab. \ref{tab:Effectiveness-of-each} shows the average time we needed for each step. Strictly speaking, we can not compare with ``previous methods'', since we are the first work on poster generation and there is no existing directly comparable work. Nevertheless, we argue that the total running time is significantly less than
the time people require to design a good poster,
it is also less than the time spent to generate the posters made by three novices in Quantitative evaluation section.\par

\begin{figure*}[thb!]
\centering
\begin{subfigure}[b]{0.28\textwidth}
\includegraphics[width=\textwidth]{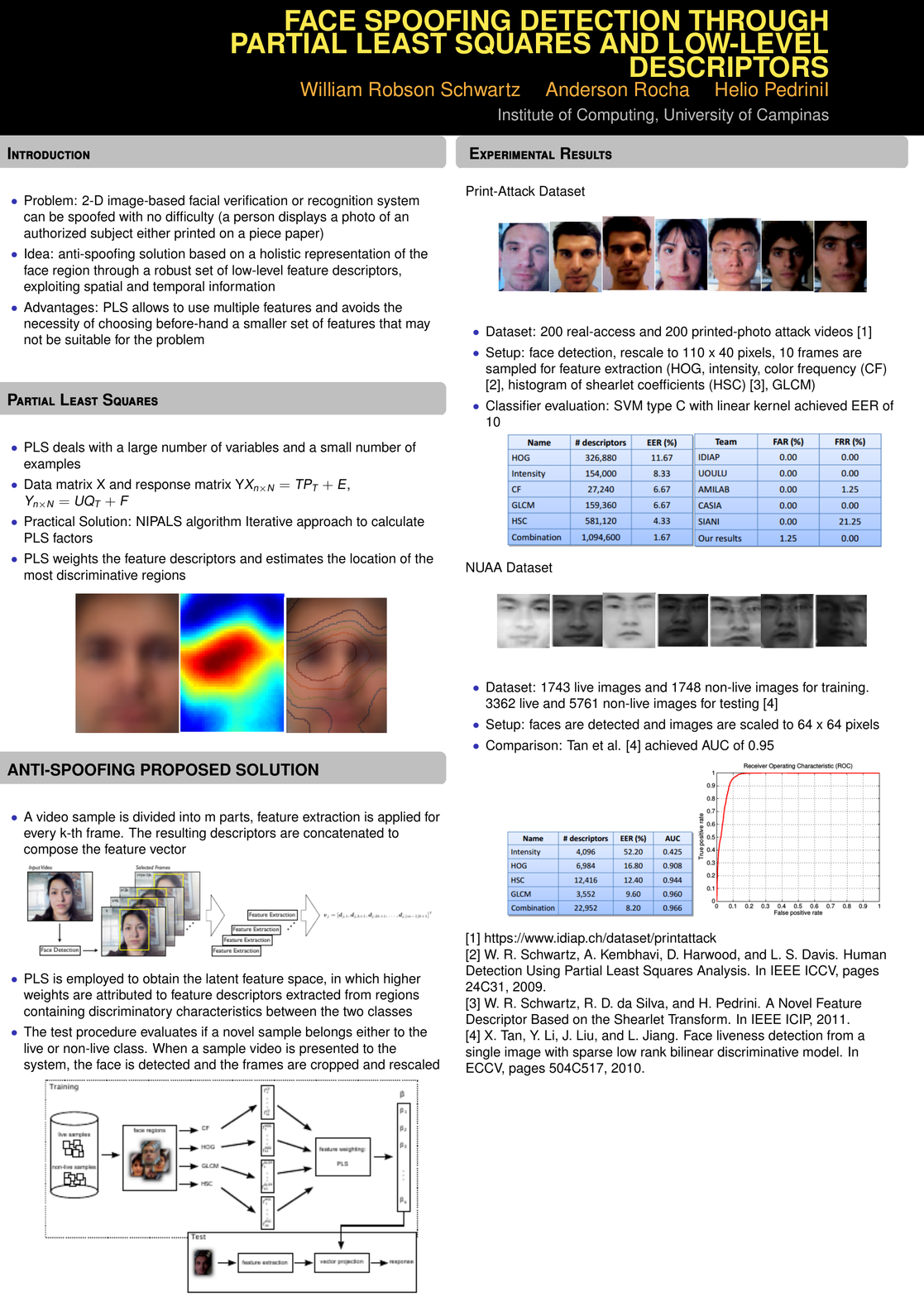}
\caption{\label{fig:novice_res}Designed by novice}
\end{subfigure}
\begin{subfigure}[b]{0.28\textwidth}
\includegraphics[width=\textwidth]{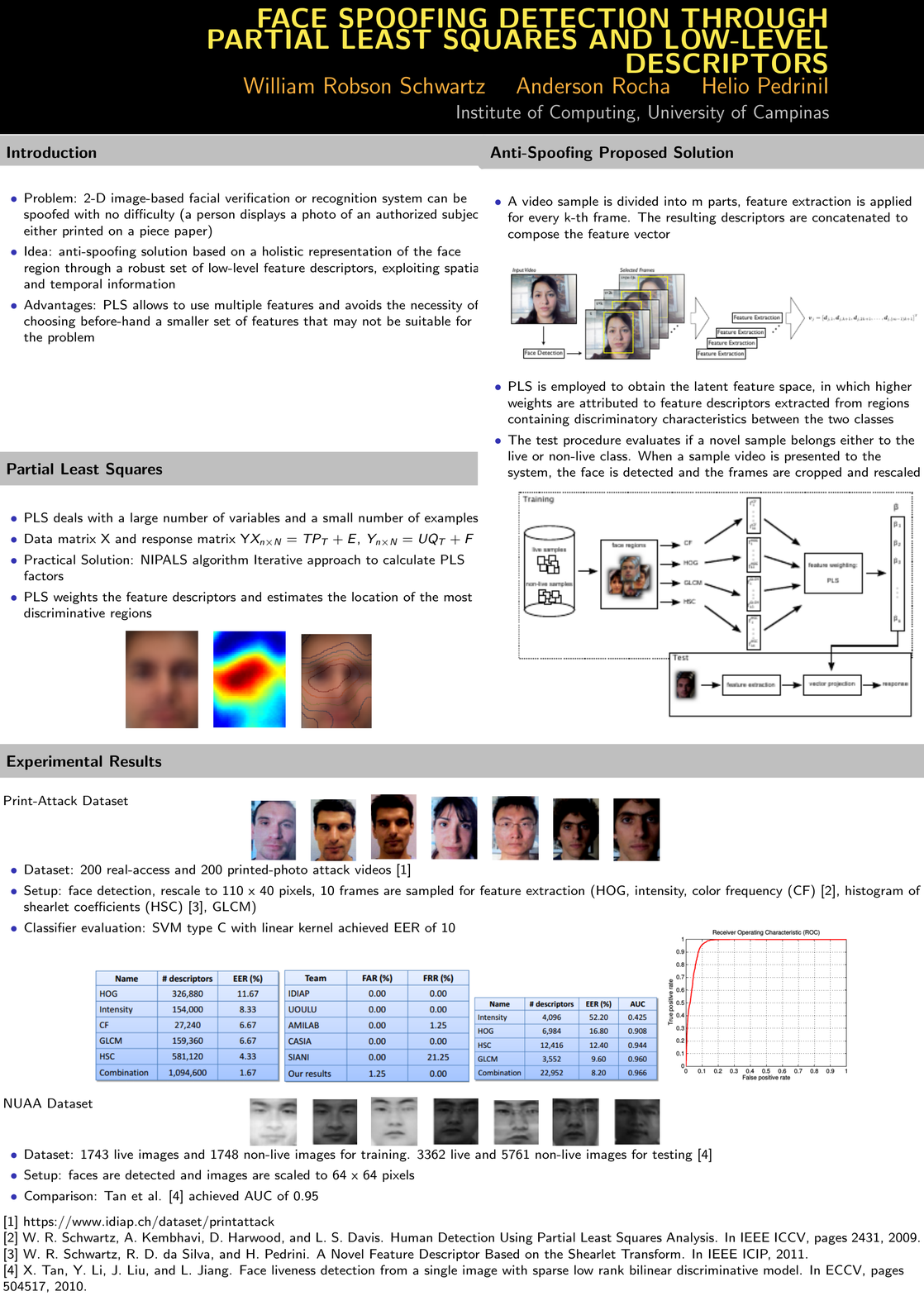}
\caption{\label{fig:our_res}Our result}
\end{subfigure}
\vspace{1em}
\begin{subfigure}[b]{0.33\textwidth}
\includegraphics[width=\textwidth]{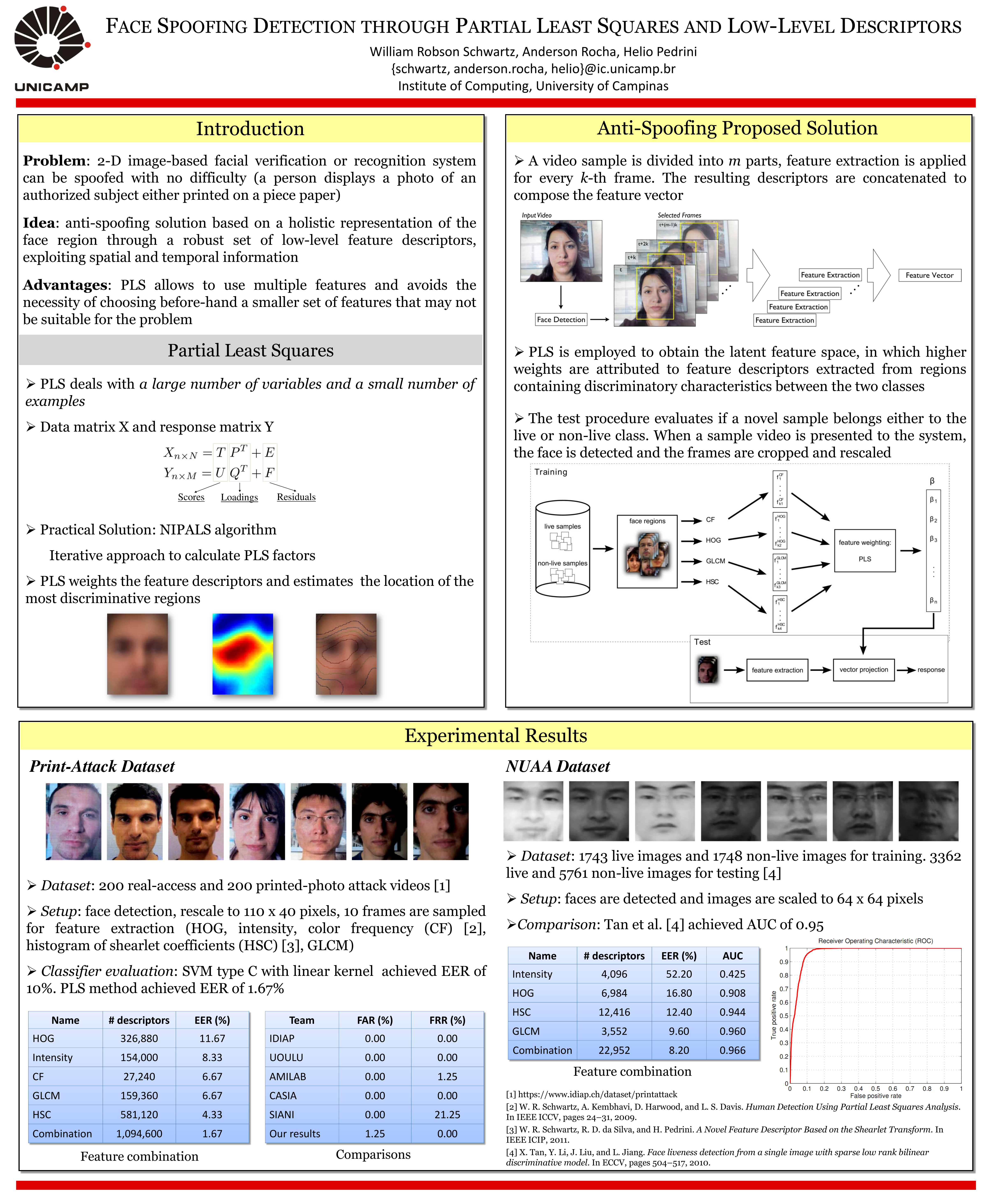}
\caption{\label{fig:original_res}Original poster}
\end{subfigure}
\caption{\label{fig:results}Results generated by different ways}
\label{fig:results}
\end{figure*}

\begin{table}
\centering{}{\small{}}%
\begin{tabular}{|c|c|r|}
\hline
\multicolumn{2}{|c|}{\textbf{\small{}stage}} & \textbf{\small{}Average time}{\small{} }\tabularnewline
\hline
\multicolumn{2}{|c|}{{\small{}Text extraction$\star$}} & {\small{}28.81s }\tabularnewline
\hline
\multirow{2}{*}{{\small{}Panel attributes inference }} & {\small{}learn } & {\small{}0.85s }\tabularnewline
\cline{2-3}
 & {\small{}infer } & {\small{}0.013s }\tabularnewline
\hline
\multicolumn{2}{|c|}{{\small{}Panel layout generation}} & {\small{}0.13s }\tabularnewline
\hline
\multirow{2}{*}{{\small{}Composition within panel}} & {\small{}learn } & {\small{}2.17s }\tabularnewline
\cline{2-3}
 & {\small{}infer } & {\small{}\small{0.03s+19.09s$\star$} }\tabularnewline
\hline
\end{tabular}\protect\caption{\label{tab:Effectiveness-of-each}Running time of each step. $\star$: it takes us 0.03s  for inference computation and the 19.09s time for latex file generation. }
\end{table}

\noindent\textbf{Quantitative Evaluation.} We quantitatively evaluate the effectiveness
of our approach.

\textbf{(1) Effectiveness of panel inference.} For this
step, we compare the inferred size and aspect ratio of panels with
the trivial solution -- SVR which trains a linear regressor%
\footnote{$s_{p}$ and $r_{p}$ are used as features for SVR. The parameters
are chosen using cross-validation. Nonlinear kernels (such as RBF) perform worse
due to over-fitting on training data. %
} to predict the panel size and panel aspect ratio from training data.
We use the panel attributes from the original posters%
\footnote{Note that, though the panels of original poster may not be the best
ones, they are the best candidate to serve as the ground truth here. %
} as the ground-truth and compute the mean-square error (MSE) of inferred
values versus ground-truth values. Our results can achieve 3650.4
and 0.67 for panel size and aspect ratio. By contrast, the values
of SVR method are 3831.3 and 0.76 respectively. This shows that our
algorithm can better estimates the panel attributes than SVR.

\begin{table*}
\begin{centering}
\begin{tabular}{|c|c|c|c|c|}
\hline
Metric  & Readability  & Informativeness  & Aesthetics  & Avg.\tabularnewline
\hline
\hline
Our method & 6.94  & 7.06 & 6.86 & 6.95\tabularnewline
\hline
Posters by novices  & 6.69  & 6.83  & 6.12  & 6.54\tabularnewline
\hline
Original posters & 7.08  & 7.03  & 7.43  & 7.18\tabularnewline
\hline
\end{tabular}
\par\end{centering}

\protect\caption{\label{tab:User-Study-of}User study of different posters generated.}
\end{table*}

\textbf{(2) User study.} User study is employed to compare
our results with original posters and posters made by novices. We
invited 10 researchers (who are experts on the evaluated topic
and kept unknown to our projects) to evaluate these results on
readability, informativeness and aesthetics. Each researcher is sequentially
shown the three results generated (in randomized order) and asked to score
the results from $0-10$, where $0$, $5$ and $10$ indicate
the lowest, middle and highest scores of corresponding metrics. The final
results are averaged for each metric item.

As in Tab. \ref{tab:User-Study-of}, our method is comparable
to original posters on \emph{readability} and \emph{informativeness};
and it is significantly better than posters made by novices. This validates
the effectiveness of our method, since the inferred panel attributes and
generated panel layout will save most valuable and important information.
In contrast, our method is lower than the original posters on aesthetics
metric (yet, still higher than those from novices). This is reasonable,
since aesthetics is a relatively subjective metric and
aesthetics generally requires a ``human touch".
It is an open problem to generate more aesthetic posters from papers.

\begin{figure}
\centering
\begin{subfigure}[t]{0.23\textwidth}
\includegraphics[width=\textwidth]{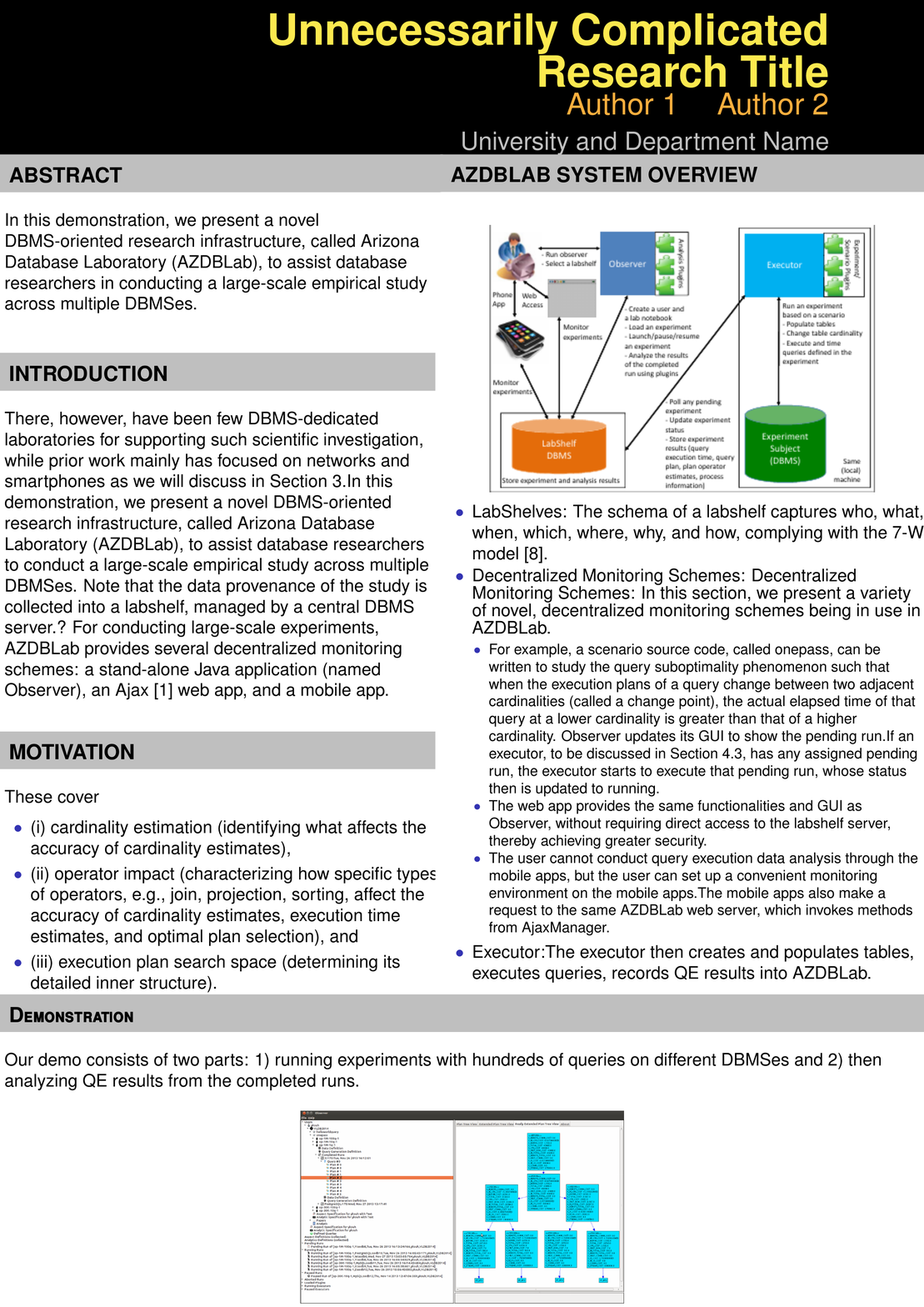}
\protect{\caption{\label{fig:resa}}}
\end{subfigure}
\begin{subfigure}[t]{0.23\textwidth}
\includegraphics[width=\textwidth]{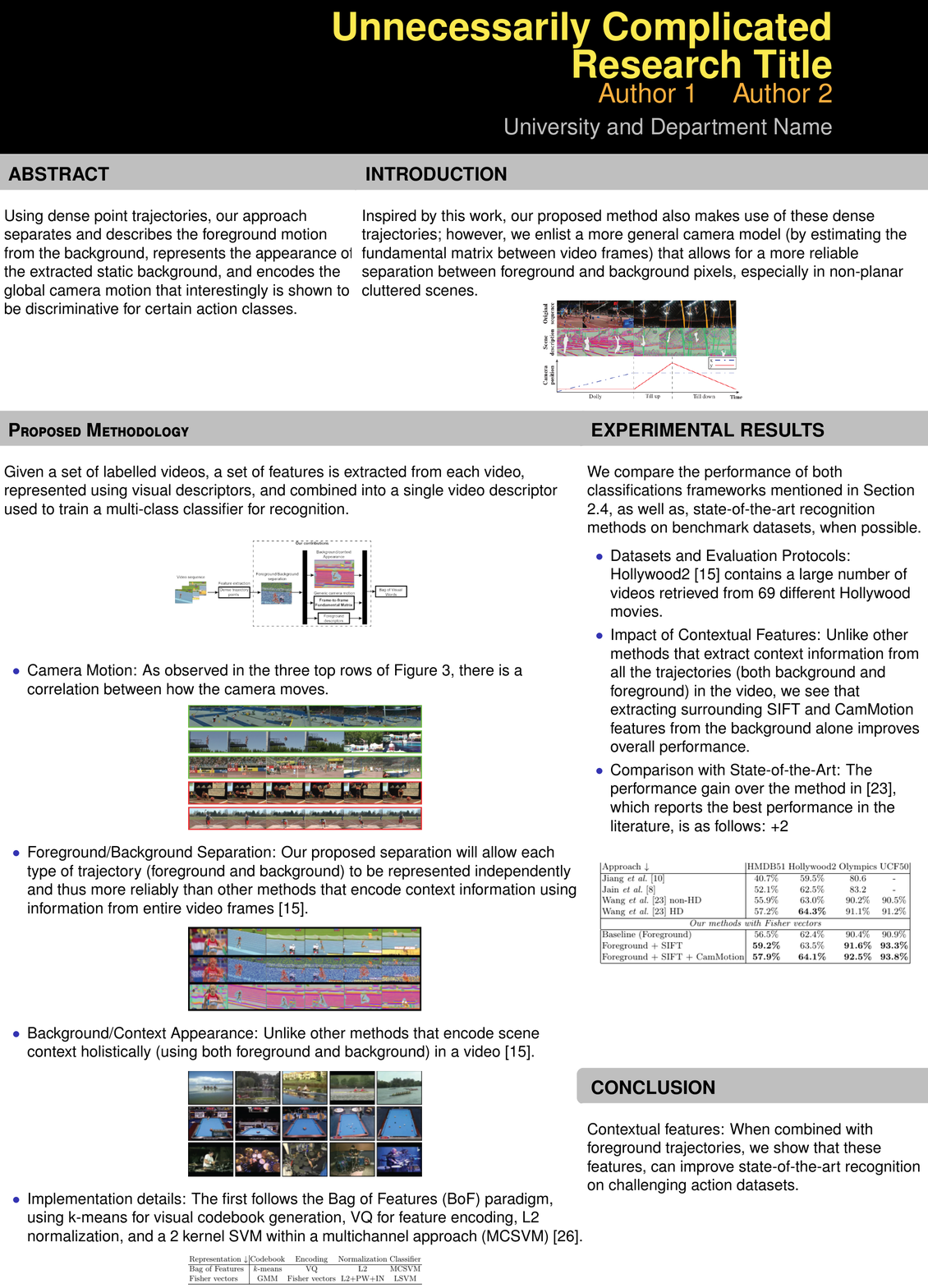}
\caption{ \label{fig:resb}}
\end{subfigure}
\protect\caption{\label{fig:res} Qualitative comparison of our result (b) and novice's result (a). Please refer to supplementary material for larger size figures.}
\end{figure}

\noindent\textbf{Qualitative Evaluation of Three Methods. }We qualitatively
compare our result (Figure \ref{fig:results}(b)) with the poster
from novices in Figure \ref{fig:results}(a) and the original poster Figure \ref{fig:results}(c). All of them are for the same paper.

It is interesting to show that if compared with the panel layout
of original poster, our panel layout looks more similar to the original one than the one by
novices. This is due to, first, the Poster-Paper dataset has a relatively
similar graphical design with high quality, and second, our split and panel
layout algorithms that work well to simulate the way how people design posters.
In contrast, the poster designed by novices in Figure \ref{fig:results}(a) has two columns, which appears less attractive to our 10 researchers; it takes the novices around 2 hours to finish all the posters.

\noindent \textbf{Further Qualitative Evaluation.} We further qualitatively evaluate our results (Figure \ref{fig:res}) by the general graphical
design principles \cite{odonovan2014}, i.e., \emph{flow}, \emph{alignment},and \emph{overlap and boundaries}.

\textbf{Flow} It is essential for a scientific poster to present information in
a clear read-order, i.e. readability. People always read a scientific
poster from left to right and from top to bottom. Since Algorithm \ref{alg:panel_gen}
recursively splits the page of poster into \emph{left, right} or \emph{top, bottom}, the panel layout we generate ensure that the read-order
matches the section order of original paper. Within each panel, our
algorithm also sequentially organizes contents which also follow
the section order of original paper and this improves the readability.

\textbf{Alignment}. Compared with the complex
alignment constraint in \cite{odonovan2014}, our formulation is much simpler and uses an enumeration variable
to indicate the horizontal position of graphical elements $h_{g}$. 
 This simplification
does not spoil our results which still have reasonable alignment as
illustrated in Figure \ref{fig:res} and quantitatively evaluated
by three metrics in Tab. \ref{tab:User-Study-of}.

\textbf{Overlap and boundaries}. Overlapped panels will
make the poster less readable and less esthetic. To avoid
this, our approach (1) recursively splits the page for panel layout;
(2) sequentially arranges the panels; (3) enforces the constraint
Eq. \ref{eq:constraint} to penalize the cases of overlapping
between graphical elements and panel boundaries. As a result, our algorithm
can achieve reasonable results without significant overlapping and/or
crossing boundaries. Similar to the manually created poster
-- Figure \ref{fig:results}(c), our result (i.e., Figure \ref{fig:results}(b))
does not have significantly overlapped panels and/or boundaries.

\noindent

\section{Conclusion and Future Work}

Automatic tools for scientific poster generation are important for
poster designers. Designers can save a lot of time with these kinds of tools.
Design is a hard work, especially for scientific posters, which
require careful consideration of both utility and aesthetics.
Abstract principles about
scientific poster design can not help designers directly. By contrast,
we propose an approach to learn design patterns from existing examples,
and this approach will hopefully lead to an automatic tool for scientific poster
generation to aid designers.

Except for scientific poster design, our approach also provides a
framework to learn other kinds of design patterns, for example web-page
design, single-page graphical design and so on. And by providing different
set of training data, our approach could generate different layout styles. Our work has several limitations. We do not consider font types in our
current implementation and only adopt a simple yet effective aesthetic metric. We plan to address these problems in future.

\section{Acknowledgements}
We would like to thank the anonymous reviewers for their insightful suggestions in improving this paper.

 \bibliographystyle{aaai}
\bibliography{reference}

\end{document}